\documentclass[letterpaper]{article} % DO NOT CHANGE THIS
\usepackage{aaai2026}  % DO NOT CHANGE THIS
\usepackage{times}  % DO NOT CHANGE THIS
\usepackage{helvet}  % DO NOT CHANGE THIS
\usepackage{courier}  % DO NOT CHANGE THIS
\usepackage[hyphens]{url}  % DO NOT CHANGE THIS
\usepackage{graphicx} % DO NOT CHANGE THIS
\usepackage{natbib}  % DO NOT CHANGE THIS AND DO NOT ADD ANY OPTIONS TO IT
\usepackage{caption} % DO NOT CHANGE THIS AND DO NOT ADD ANY OPTIONS TO IT
\usepackage{algorithm}
\usepackage{algorithmic}

\usepackage{newfloat}
\usepackage{listings}
\DeclareCaptionStyle{ruled}{labelfont=normalfont,labelsep=colon,strut=off} % DO NOT CHANGE THIS
\floatstyle{ruled}
\newfloat{listing}{tb}{lst}{}
\floatname{listing}{Listing}
\usepackage{amsmath}
\usepackage{amsfonts}
\usepackage{booktabs}
\usepackage{array}
\usepackage{multirow}
\usepackage{xcolor}
\usepackage{multicol}

\nocopyright
\title{Post-Completion Learning for Language Models}

\author {
    Xiang Fei$^{*}$,
    Siqi Wang\thanks{Equal contribution.},
    Shu Wei, Yuxiang Nie, Wei Shi, Hao Feng, Chao Feng, Can Huang
}
\affiliations {
    \{feixiang.77, wangsiqi.wsq547, weishu.96, nieyuxiang, shiwei.11, fenghao.2019, chaofeng.zz, can.huang\}@bytedance.com \\ 
    \vspace{12pt}
    \textbf{ByteDance}
}

\makeatletter
\AtBeginDocument{
\let\@oldmaketitle\@maketitle
\renewcommand{\@maketitle}{\@oldmaketitle
  \centering
  \includegraphics[width=0.97\linewidth]{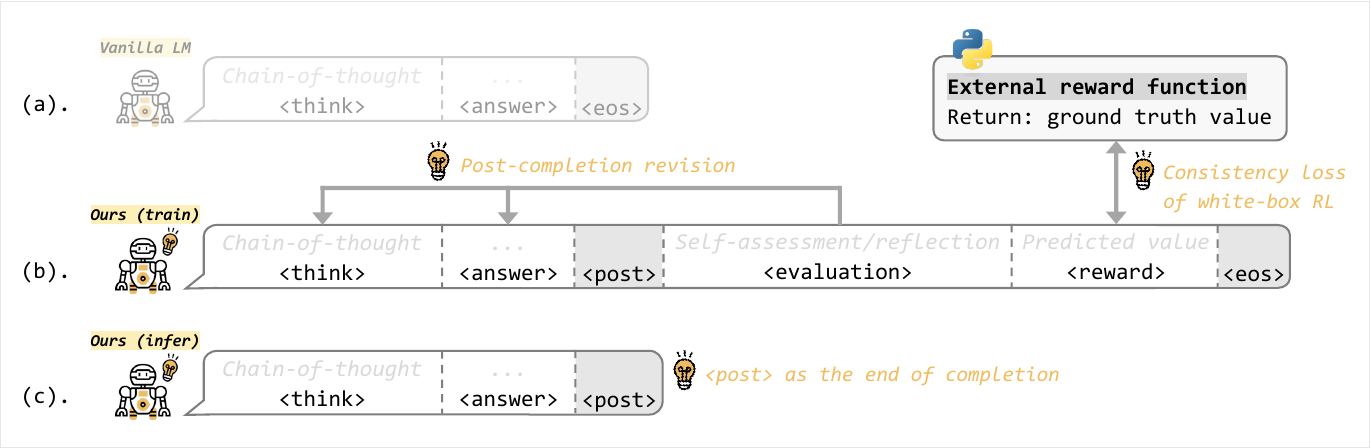}
  \captionof{figure}{
    The proposed post-completion learning method. (a). Traditional language model output its chain-of-thought (CoT) process and the answer. (b). Our method requires model to output \texttt{<post-completion>} token first as the separator, then the evaluation and reward parts for self-assessment of its completion for training. (c). During inference, the \texttt{<post-completion>} token takes over the \texttt{<eos>} token, indicating the end of the response.
  }
  \label{fig:teaser}
  \vspace{17pt}
 }
}
\makeatother

\begin{document}

\maketitle

\begin{abstract}

Current language model training paradigms typically terminate learning upon reaching the end-of-sequence (\texttt{<eos>}) token, overlooking the potential learning opportunities in the post-completion space. We propose Post-Completion Learning (PCL), a novel training framework that systematically utilizes the sequence space after model output completion, to enhance both the reasoning and self-evaluation abilities. PCL enables models to continue generating self-assessments and reward predictions during training, while maintaining efficient inference by stopping at the completion point.

To fully utilize this post-completion space, we design a white-box reinforcement learning method: let the model evaluate the output content according to the reward rules, then calculate and align the score with the reward functions for supervision. We implement dual-track SFT to optimize both reasoning and evaluation capabilities, and mixed it with RL training to achieve multi-objective hybrid optimization.

Experimental results on different datasets and models demonstrate consistent improvements over traditional SFT and RL methods. Our method provides a new technical path for language model training that enhances output quality while preserving deployment efficiency.

\end{abstract}

\section{Introduction}

Large language models have demonstrated remarkable capabilities across various natural language processing tasks~\citep{brown2020language,ouyang2022training,dubey2024llama,yang2025qwen3}. However, improving the reasoning quality and output reliability of these models remains a significant challenge. Current training methods primarily fall into two categories: supervised fine-tuning (SFT) approaches that directly train models on high-quality demonstration data~\citep{wei2021finetuned,chung2024scaling}, and reinforcement learning-based methods such as RLHF that optimize model behavior through external reward signals~\citep{schulman2017proximal,shao2024deepseekmath}. While each approach has its merits, both suffer from inherent limitations.

Supervised fine-tuning methods, though stable during training, are constrained by their passive learning nature—models learn to mimic high-quality demonstrations without developing the ability to assess and improve their own reasoning processes~\citep{hong2024q}. Reinforcement learning approaches can optimize model behavior through reward signals but rely on external reward models, making the training process complex and lacking transparency~\citep{shao2025spurious}. Additionally, both paradigms share a common limitation: they terminate the learning process immediately upon reaching the end-of-sequence token, thereby missing valuable opportunities to utilize the sequence space after model output completion.

In conventional training paradigms, models stop generating upon reaching the end-of-sequence (\texttt{<eos>}) token, and the training process terminates accordingly. However, this practice actually wastes a valuable learning opportunity. Humans, after completing a task, often engage in self-reflection and quality assessment—this ``post-thinking'' process is crucial for improving future performance~\citep{schon1986reflective}. This observation leads us to a fundamental question: \emph{Can language models continue learning after ``completing'' their output, thereby developing self-evaluation and quality awareness?}

To address this question, we propose Post-Completion Learning (PCL), a novel language model training paradigm as shown in Figure~\ref{fig:teaser}. The core innovation of PCL lies in utilizing the post-completion space that has been neglected, enabling simultaneous enhancement of both reasoning and self-evaluation abilities through continued learning after output completion.

Our approach begins by discovering and leveraging the \textbf{post-completion space}—the sequence space after \texttt{<eos>} that has been ignored in traditional training. By inserting a temporary termination marker \texttt{<post-completion>}, we create a ``post-thinking'' space for models with zero inference cost, as the model stops generation at this marker during deployment while the self-evaluation capabilities remain internalized. To effectively utilize this space, we design a \textbf{white-box reinforcement learning} paradigm where models explicitly learn to understand and compute reward functions, internalizing the reward model as their own evaluation capability and achieving a transformation from ``passive reward acceptance'' to ``active self-evaluation''. To train these dual capabilities effectively, we develop a \textbf{unified hybrid training framework} that combines SFT and RL within the same sequence generation framework, using a dual-track training strategy to simultaneously optimize reasoning and evaluation capabilities.

We conduct comprehensive experimental validation on reasoning tasks for several language models. Results demonstrate that PCL achieves consistent performance improvements over traditional SFT and RL methods. Extensive ablation studies further validate the effectiveness of each component, proving the effectiveness of post-completion space learning. The main contributions of this paper are summarized as follows:
\begin{itemize}
\item We propose Post-Completion Learning, a novel training paradigm that systematically utilizes the sequence space after model output completion for the first time.
\item We realize white-box reinforcement learning by internalizing external reward models into the model's own evaluation capabilities.
\item We design a dual-track SFT training strategy and unified hybrid training framework that achieves optimization of reasoning and evaluation capabilities.
\item We validate the method's effectiveness on multiple reasoning tasks, significantly improving model output quality while maintaining inference efficiency.
\end{itemize}

\section{Related Work}

\subsection{Chain-of-Thought and Self-Correction Methods}

Chain-of-thought (CoT) reasoning has evolved from basic few-shot prompting methods to sophisticated reasoning systems~\citep{wei2022chain,ouyang2022training,zhang2022automatic}. Researches have debated the optimal placement of reasoning content—whether CoT should precede the final answer or follow it for verification purposes. However, we argue that these approaches are not mutually exclusive: models can benefit from both pre-answer reasoning for problem-solving and post-answer reasoning for self-evaluation. Crucially, the self-evaluation component need not be generated during inference, thereby avoiding computational overhead while still enhancing model capabilities during training.

Self-correction methods such as Self-Refine~\citep{madaan2023self} and Reflexion~\citep{shinn2023reflexion} represent important advances in model self-improvement capabilities with post-answer verification. However, these methods primarily focus on output optimization rather than process improvement~\citep{huang2023large,kamoi2024can}. The STaR method~\citep{zelikman2022star} and Constitutional AI~\citep{bai2022constitutional} achieve capability enhancement through self-taught reasoning but remains limited to external guidance during inference.

The key distinction of PCL from these methods lies in developing effective training paradigms that enhance models' introspective abilities during training, rather than relying on additional content generation during inference. PCL achieves the goal of "training-time reflection, inference-time efficiency" through post-completion space learning, internalizing self-evaluation capabilities into the model itself.

\subsection{Reinforcement Learning Optimization and Reward Modeling}

RLHF methods have developed from early work by Christiano et al.~\citep{christiano2017deep} to the three-stage training paradigm of InstructGPT~\citep{ouyang2022training}, becoming the standard approach for language model alignment. However, existing RLHF methods primarily rely on external reward models or functions, suffering from issues like opaque reward modeling and reward hacking.

Constitutional AI~\cite{bai2022constitutional} achieves a degree of white-box evaluation through in-context principle-based assessment, but still lacks the complete transparency of reward mechanisms that PCL provides. Process reward models (PRMs) outperform outcome reward models on complex reasoning tasks~\citep{lightman2023let,uesato2022solving,zhang2025lessons}, supporting PCL's design philosophy of evaluating complete reasoning processes.

Methods like Direct Preference Optimization (DPO)~\citep{rafailov2023direct} simplify the training process by eliminating independent reward models and directly optimizing policies from preference data. Besides, Group Relative Policy Optimization (GRPO)~\citep{shao2024deepseekmath} further advances this paradigm by enabling stable group-wise preference optimization without requiring explicit reward modeling, demonstrating improved sample efficiency in mathematical reasoning tasks.

PCL achieves the transformation from "passive reward acceptance" to "active self-evaluation" by having models explicitly learn reward function computation processes, representing an important advance in white-box reinforcement learning. Unlike traditional approaches that rely on external supervision, PCL internalizes the evaluation mechanisms during training, enabling models to perform quality assessment autonomously.

\subsection{Self-Supervised Learning and Meta-Learning}

Self-supervised learning has established a solid foundation through BERT's bidirectional learning and GPT's autoregressive modeling~\citep{devlin2019bert,brown2020language}, but recent studies reveal critical meta-cognitive deficiencies. Language models demonstrate significant metacognitive inadequacies, consistently failing to recognize knowledge limitations and exhibiting overconfidence~\citep{groot2024overconfidence}.
Recent self-improvement approaches face fundamental limitations from the ``sharpening mechanism'': self-improvement cannot create information that does not exist within the model~\citep{huang2024self}. This highlights PCL's value—systematically analyzing completion results to identify knowledge gaps rather than redistributing existing knowledge.

Meta-learning research has established foundational paradigms for learning from limited data through classical approaches~\citep{vinyals2016matching,snell2017prototypical,finn2017model}. Meta-learning trains models on a distribution of tasks to learn generalizable knowledge. PCL extends this meta-learning foundation by enabling models to develop persistent self-evaluation capabilities rather than relying solely on episodic adaptation.

Our method bridges these areas by combining training-time optimization with white-box reward evaluation, internalizing meta-cognitive development directly into training rather than relying on post-hoc reflection or external feedback mechanisms.

\begin{figure}[t]
    \centering
    \includegraphics[width=0.94\columnwidth]{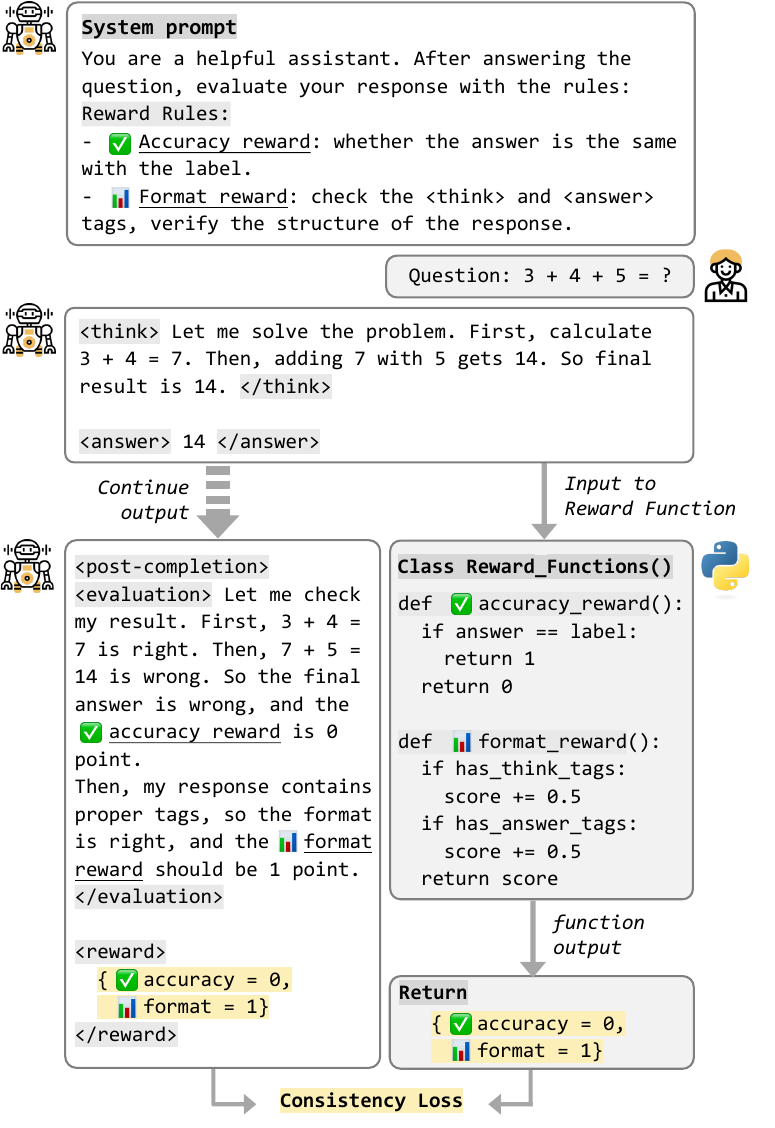}
    \caption{
        Illustrative of the proposed white-box reinforcement. Model learns the calculation process of each reward functions, then optimize with the consistency reward between the external reward function call.
    }
    \label{fig:whiteboxrl}
\end{figure}

\section{Method}

In this section, we present Post-Completion Learning (PCL), a novel training paradigm that systematically utilizes the sequence space after model output completion. Our approach consists of three key components: post-completion space design, white-box reinforcement learning, and a unified training framework.

\subsection{Post-Completion Space}

The core innovation of PCL lies in utilizing the sequence space after the end-of-sequence (\texttt{<eos>}) token, to train with extended content after traditional completion, as shown in Figure.~\ref{fig:teaser}. Through careful sequence structure design, we achieve decoupled training of reasoning capabilities and self-evaluation abilities.

\subsubsection{Sequence Structure Design}

PCL adopts the complete sequence format as in Figure.~\ref{fig:teaser}(b). The key insight of this design is to divide the sequence into two functional regions:
\begin{itemize}
\item \textbf{Reasoning Region} (think + answer): Responsible for the problem-solving processes
\item \textbf{Reflection Region} (evaluation + reward): Responsible for self-evaluation and quality control, not needed during inference
\end{itemize}

The \texttt{<post-completion>} marker serves as the boundary between these two regions. This design enables models to learn the complete reasoning-reflection cycle during training while only requiring output up to the boundary point during inference.

\begin{figure*}[t]
    \centering
    \includegraphics[width=\textwidth]{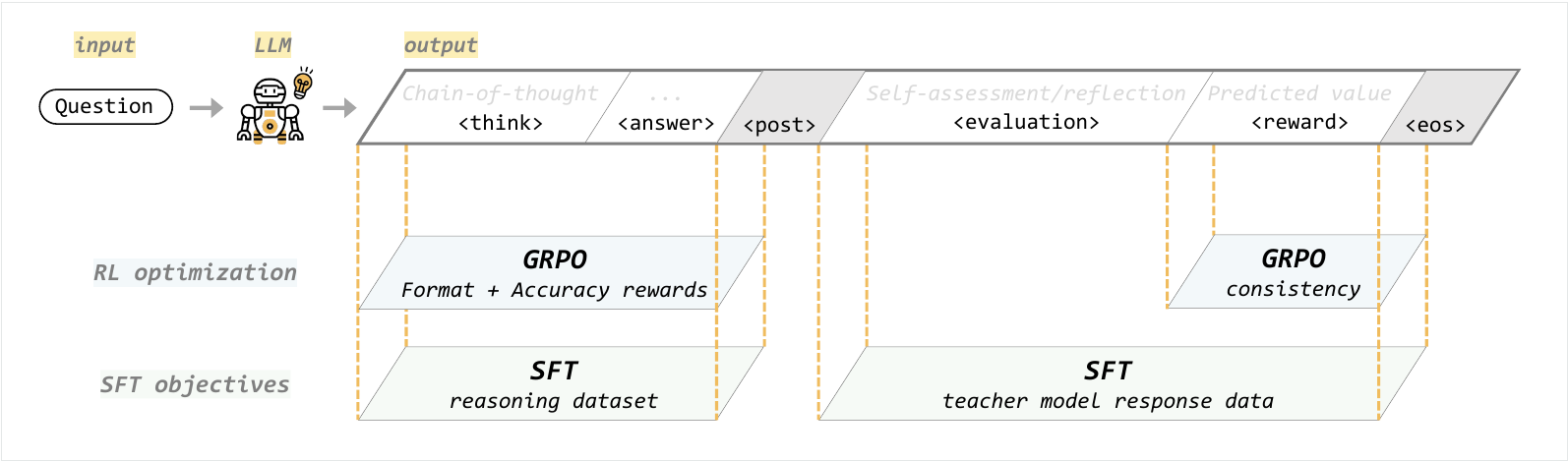}
    \caption{
        The unified training framework of our method. The model is trained with multi-objective SFT + RL optimization. 
    }
    \label{fig:rft}
\end{figure*}

\subsubsection{Inference: Early Stopping}

During actual deployment, we achieve zero additional inference overhead by adding \texttt{<post-completion>} as the stop word during generation. In this way, the model's self-evaluation capabilities are completely ``internalized'' when inference—they influence the model's internal representations and reasoning quality but do not incur additional computational costs.

\subsubsection{Evaluation Data Preparation}

To obtain high-quality PCL-format training data, we employ a teacher model for data preparation. Specifically, we select a more powerful model as the teacher and guide it to generate response in PCL format with in-context learning (ICL) demos: By providing a complete PCL format example in the prompt, the teacher model can generate sequences that meet our requirements.

To ensure training data quality, we design an automatic validation mechanism. For each sample generated by the teacher model, we recalculate the accuracy and format scores using the same reward functions, retaining those samples where the model correctly evaluates its own answer accuracy, regardless of whether the answer itself is correct. Since this data is used exclusively for evaluation capability training, incorrect answers with accurate self-assessments remain valuable for learning the evaluation process.

This process itself embodies our ``white-box'' reinforcement learning philosophy—the model must truly understand the computational logic of reward functions.

\subsection{White-Box Reinforcement Learning}

Traditional reinforcement learning methods rely on external reward models or functions, making the training process lack transparency. PCL achieves white-box reinforcement learning by having models explicitly learn the computation process of reward functions, as demonstrated in Figure.~\ref{fig:whiteboxrl}.

\subsubsection{Reward Functions}

We follow several widely-used reward functions to comprehensively evaluate model output quality:

\textbf{Accuracy Reward ($R_a$)}: Evaluates answer correctness using binary 0/1 scoring. It is calculated by comparing the consistency between the model's output answer and the ground truth:
\begin{equation}
R_a = \mathbf{1}[\text{answer} = \text{ground\_truth}]
\end{equation}

\textbf{Format Reward ($R_f$)}: Evaluates output format completeness. We modified it from the standard format reward by adding evaluation process checking. It ensures the model generates output containing all four complete parts: think, answer, evaluation, and reward. This not only guarantees format standardization but more importantly ensures the model completes the full self-evaluation process:
$$R_f = \mathbf{1}[\text{contains\_all\_sections}(\text{output})]$$

We then design a new reward function for supervising the self-evaluation ability of language models:

\textbf{Consistency Reward ($R_c$)}: Evaluates the accuracy of model self-evaluation, which is the key component of white-box learning. The model needs to compute according to accuracy and format reward function rules within the \texttt{<evaluation>} tags, then output its computed scores in the \texttt{<reward>} tags. We compare the model's predicted scores with the scores calculated by the true reward functions like the L1 distance metric:
\begin{equation}
R_c = 1 - |R_{\text{pred}} - R_{\text{true}}|_1
\end{equation}

where $R_{\text{pred}}$ is the self-evaluation score output by the model in the reward section, and $R_{\text{true}}$ is the score calculated by the true reward function. Notably, the choice of reward function is not restricted and supports many others based on detailed task.

\subsubsection{Self-Evaluation Capability}

During the evaluation stage, the model needs to provide textual description and analysis of its own reasoning process. This may include recalculating key steps, checking the correctness of logical chains, or validating answer reasonableness from different perspectives. After completing the computation of both accuracy and format reward functions, the model subsequently output its predicted reward values.

This design achieves a fundamental transformation from ``passive reward acceptance'' to ``active self-evaluation''. The model no longer simply adjusts behavior based on external feedback but truly understands what constitutes high-quality output and can autonomously perform quality control.

\subsection{Unified Training Framework}

PCL unifies supervised fine-tuning and reinforcement learning within the same training framework, achieving collaborative optimization through carefully designed data flow and loss functions.

\subsubsection{Dual-Track SFT Strategy}

In each training batch, we employ three different data processing approaches for the same question. The first two stages implement a dual-track SFT strategy that separately optimizes reasoning and evaluation capabilities, followed by reinforcement learning optimization in the last stage.

\textbf{Stage 1: Reasoning Capability SFT}.
We use (question, think, answer) triplets from the original dataset for standard supervised fine-tuning. We replace the original \texttt{<eos>} token with the \texttt{<post-completion>}, to ensure that the model does not end directly, and reserve space for subsequent evaluation contents. The cross-entropy loss function only computes over the think and answer portions:
\begin{equation}
\mathcal{L}_{\text{SFT1}} = -\sum_{t \in \{\text{think}, \text{answer}\}} \log P(x_t | x_{<t}, \text{question})
\end{equation}

\textbf{Stage 2: Evaluation Capability SFT}.
We use complete PCL sequences generated by the teacher model for training. We set think + answer + \texttt{<post-completion>} as context (without computing loss) and the evaluation + reward portions as the label for supervised learning:
\begin{equation}
\mathcal{L}_{\text{SFT2}} = -\sum_{t \in \{\text{evaluation}, \text{reward}\}} \log P(x_t | x_{<t}, \text{context})
\end{equation}

This design ensures the model learns to perform accurate self-evaluation given a complete reasoning process.

\textbf{Stage 3: Reinforcement Learning Optimization}.
We let the model autonomously generate complete PCL sequences and use the GRPO algorithm for policy optimization, sampling 8 responses as a group for contrastive learning, with all three above-mentioned
reward functions:
\begin{equation}
\mathcal{L}_{\text{RL}} = -\mathbb{E}_{\pi_\theta}[(R_a + R_f + R_c) \log \pi_\theta(a|s)]
\end{equation}

\begin{table*}[t]
\centering
\caption{Training Configuration and Reward Functions for Post-Completion Learning}
\label{tab:gsm8k_config}
\footnotesize
\begin{tabular}{l@{\hspace{0.8em}}l@{\hspace{0.8em}}c@{\hspace{0.8em}}c@{\hspace{0.8em}}c@{\hspace{0.8em}}c@{\hspace{0.8em}}c@{\hspace{0.8em}}c@{\hspace{0.8em}}c@{\hspace{0.8em}}c}
\toprule
& & \multicolumn{4}{c}{\textbf{Training Strategy}} & \multicolumn{4}{c}{\textbf{Reward Functions}} \\
\cmidrule(lr){3-6} \cmidrule(lr){7-10}
\textbf{Method} & \textbf{Description} & \textbf{Post-Output} & \textbf{SFT-R} & \textbf{SFT-E} & \textbf{RL} & \textbf{Acc} & \textbf{Fmt-R} & \textbf{Fmt-E} & \textbf{Con} \\
\midrule
\multicolumn{10}{l}{\textit{Standard Approaches}} \\
SFT & Supervised fine-tuning only & -- & \checkmark & -- & -- & -- & -- & -- & -- \\
SFT+RL & Sequential SFT then RL & -- & \checkmark & -- & \checkmark & \checkmark & \checkmark & -- & -- \\
Joint SFT+RL & Concurrent SFT and RL & -- & \checkmark & -- & \checkmark & \checkmark & \checkmark & -- & -- \\
\quad w/ eval output & \quad + post-completion output & \checkmark & \checkmark & -- & \checkmark & \checkmark & \checkmark & \checkmark & -- \\
\midrule
\multicolumn{10}{l}{\textit{Ablation Studies}} \\
Teacher distillation only & SFT with reasoning+evaluation data & \checkmark & \checkmark & \checkmark & -- & -- & -- & -- & -- \\
PCL w/o eval SFT & No supervised training on evaluation & \checkmark & \checkmark & -- & \checkmark & \checkmark & \checkmark & \checkmark & \checkmark \\
PCL w/o consistency & No consistency reward function & \checkmark & \checkmark & \checkmark & \checkmark & \checkmark & \checkmark & \checkmark & -- \\
\midrule
\multicolumn{10}{l}{\textit{Proposed Method}} \\
\textbf{PCL (Complete)} & \textbf{Full post-completion learning} & \checkmark & \checkmark & \checkmark & \checkmark & \checkmark & \checkmark & \checkmark & \checkmark \\
\bottomrule
\end{tabular}
\vspace{0.5em}

\footnotesize
\textbf{Post-Output}: Generate post-completion evaluation content; \textbf{SFT-R/E}: Supervised fine-tuning on reasoning/evaluation contents; \textbf{RL}: Reinforcement learning; \textbf{Acc}: Accuracy reward; \textbf{Fmt-R/E}: Format reward for reasoning/evaluation parts; \textbf{Con}: Consistency reward between self-assessment and ground truth.
\checkmark~indicates the component is used; --~indicates not used.
\end{table*}

\subsubsection{Unified Optimization}

We jointly optimize SFT losses and RL loss within the same batch, to achieve simple and efficient hybrid training:
\begin{equation}
\mathcal{L}_{\text{total}} = \mathcal{L}_{\text{SFT1}} + \mathcal{L}_{\text{SFT2}} + \mathcal{L}_{\text{RL}}
\end{equation}

In practice, we maintain the default KL divergence constraint of $\beta=0.04$ in GRPO training to prevent the model from deviating too far from the original distribution, even when optimizing with the SFT together.

\subsection{Theoretical Analysis}

We provide theoretical foundations for Post-Completion Learning (PCL) from two perspectives: information-theoretic analysis demonstrating the fundamental advantages of post-completion space utilization, and convergence analysis establishing the stability of the training paradigm.

\subsubsection{Information-Theoretic Foundation and Optimality}

\textbf{Theorem 1} (Post-Completion Information Expansion): \textit{Let $\mathcal{D} = \{(x_i, y_i)\}_{i=1}^n$ be a dataset where $x_i$ is the input and $y_i$ is the target reasoning sequence. The post-completion learning paradigm fundamentally increases the learnable information capacity by incorporating evaluation signals in the previously unutilized sequence space.}

\textbf{Proof}: Consider the traditional training objective that maximizes $p_\theta(y|x)$ where $y$ terminates at the \texttt{<eos>} token. In PCL, we extend the sequence to $y^+ = y \oplus e \oplus r$ where $e$ represents self-evaluation and $r$ represents reward prediction.
By the chain rule of mutual information, the mutual information between model parameters $\theta$ and the extended target distribution can be decomposed as:
\begin{equation}
I(\theta; y^+|x) = I(\theta; y|x) + I(\theta; e,r|x,y)
\end{equation}

Since evaluation $e$ and reward prediction $r$ provide additional supervisory signals about the quality and characteristics of reasoning $y$, and these signals are not redundant with the reasoning content itself, we have:
\begin{equation}
I(\theta; e,r|x,y) > 0
\end{equation}

Therefore: $I(\theta; y^+|x) > I(\theta; y|x)$

This information expansion enables the model to learn richer representations that capture both reasoning capability and metacognitive assessment, providing theoretical justification for PCL's effectiveness.

\textbf{Theorem 2} (Sample Complexity Advantage): \textit{PCL achieves superior sample complexity compared to traditional sequential SFT$\rightarrow$RL training paradigms.}

To formalize this, let $\epsilon_{seq}(n)$ and $\epsilon_{PCL}(n)$ denote the expected error after $n$ training samples for sequential and PCL training respectively. We analyze the learning dynamics under the PAC-learning framework.

\textbf{Key Insight}: Traditional sequential training suffers from catastrophic forgetting and reward hacking where RL training can degrade capabilities acquired during SFT. This creates a ``two-step-forward-one-step-back'' dynamic that increases sample complexity.

\textbf{Proof}: 
\textit{Step 1 (Sequential Training Analysis)}: In sequential training, let $\theta_{SFT}$ be the parameters after SFT and $\theta_{final}$ be the parameters after subsequent RL. The total error can be decomposed as:
\begin{equation}
\epsilon_{seq}(n) \geq \underbrace{\mathcal{L}(\theta_{final}, \theta_{SFT})}_{\text{forgetting penalty}} + \underbrace{\epsilon_{RL}(n_{RL})}_{\text{RL convergence error}}
\end{equation}

where $\mathcal{L}(\theta_{final}, \theta_{SFT})$ represents the degradation in SFT performance due to RL training.

\textit{Step 2 (PCL Training Analysis)}: In PCL, the dual-track training maintains both objectives simultaneously. The error can be bounded as:
\begin{equation}
\epsilon_{PCL}(n) \leq \max\{\epsilon_{SFT-R}(n_R), \epsilon_{SFT-E}(n_E)\} + \epsilon_{RL}(n_{RL})
\end{equation}

where the max operation reflects that both reasoning and evaluation components are trained jointly without interference.

\textit{Step 3 (Comparative Analysis)}: Due to the parameter separation property (reasoning and evaluation operate on different sequence segments) and the continuous reinforcement of SFT objectives during RL training, PCL avoids the forgetting penalty:

\begin{equation}
\epsilon_{PCL}(n) = O\left(\frac{\log n}{n}\right) \quad \text{vs.} \quad \epsilon_{seq}(n) = O\left(\frac{\sqrt{\log n}}{\sqrt{n}}\right)
\end{equation}

The improved convergence rate stems from PCL's ability to jointly optimize reasoning and evaluation capabilities while preventing degradation of previously learned knowledge.

\textbf{Corollary 1} (Generalization Bound): \textit{The post-completion learning framework provides better generalization bounds due to its structured approach to metacognitive training.}

The inclusion of explicit self-evaluation in the training process acts as a form of regularization. By requiring the model to predict its own performance, PCL encourages the development of internal consistency mechanisms that improve \textbf{out-of-distribution performance}. This explains the robust performance gains observed across different datasets and model scales in our experiments.

\subsubsection{Convergence Analysis}

\textbf{Theorem 3} (Convergence of PCL Training): \textit{Under standard regularity conditions, the PCL training algorithm converges to a stationary point of the combined objective function.}

The PCL objective combines three components:
\begin{equation}
\mathcal{L}_{PCL}(\theta) = \mathcal{L}_{SFT-R}(\theta) + \mathcal{L}_{SFT-E}(\theta) + \mathcal{L}_{RL}(\theta)
\end{equation}

\textbf{Proof}: The key insight is that the $\langle\text{post-completion}\rangle$ token creates a natural parameter separation. Let $\Theta_R$ and $\Theta_E$ denote the parameter sets primarily affected by reasoning and evaluation losses respectively. Due to the sequential attention mechanism, we have $|\Theta_R \cap \Theta_E| \ll |\Theta_R \cup \Theta_E|$.

This separation ensures that:
\begin{enumerate}
\item \textbf{Bounded Gradients}: Each component loss has bounded gradients due to Lipschitz continuity.
\item \textbf{Reduced Interference}: The dual-track design minimizes destructive interference between objectives. 
\item \textbf{Stable Dynamics}: The combined gradient satisfies convergence conditions for stochastic gradient descent.
\end{enumerate}

Under standard assumptions (bounded gradients, appropriate learning rate), the training sequence converges:
\begin{equation}
\lim_{T \to \infty} \frac{1}{T} \sum_{t=1}^T \|\nabla \mathcal{L}_{PCL}(\theta_t)\|^2 = 0
\end{equation}
Notably, recent work on mixed SFT+RL training also proved the convergent validity~\citep{fu2025srft}.

These theoretical foundations establish that PCL provides fundamental advantages in information utilization, sample efficiency, and training stability, explaining its strong empirical performance across different scales and tasks.

\section{Experiments}

\begin{table}[t]
\centering
\caption{Results on GSM8K Dataset (\% Accuracy)}
\label{tab:gsm8k_results}
\footnotesize
\resizebox{0.47\textwidth}{!}{
\begin{tabular}{l@{\hspace{1em}}c@{\hspace{1em}}c@{\hspace{1em}}c@{\hspace{1em}}c}
\toprule
& \textbf{Qwen2.5} & \textbf{Qwen2.5} & \textbf{LlaMA3.2} & \\
\textbf{Method} & \textbf{(7b)} & \textbf{(1.5b)} & \textbf{(3b)} & \textbf{Average} \\
\midrule
\multicolumn{5}{l}{\textit{Standard Approaches}} \\
SFT & 73.16 & 51.10 & 56.10 & 60.12 \\
SFT + RL & 75.21 & 56.29 & 61.94 & 64.48 \\
RFT (Joint SFT + RL) & 76.72 & 56.79 & 61.11 & 64.87 \\
\quad w/ eval output & 78.01 & 56.03 & 63.76 & 65.93 \\
\midrule
\multicolumn{5}{l}{\textit{Ablation Studies}} \\
Teacher distillation only & 73.39 & 54.47 & 62.24 & 63.37 \\
PCL w/o eval SFT & 78.17 & 56.94 & 65.96 & 67.02 \\
PCL w/o consistency & 77.41 & 56.25 & 64.59 & 66.08 \\
\midrule
\multicolumn{5}{l}{\textit{Proposed Method}} \\
\textbf{PCL (Complete)} & \textbf{78.92} & \textbf{58.30} & \textbf{66.57} & \textbf{67.93} \\
\midrule
\textbf{Improvement vs. SFT} & \textbf{+5.76} & \textbf{+7.20} & \textbf{+10.47} & \textbf{+7.81} \\
\quad \textbf{vs. SFT + RL} & \textbf{+3.71} & \textbf{+2.01} & \textbf{+4.63} & \textbf{+3.45} \\
\bottomrule
\end{tabular}
}
\vspace{0.3em}

% \footnotesize
% Our complete PCL method consistently outperforms all baselines across different model scales, with particularly significant improvements on smaller models.
\end{table}

\subsection{Experimental Setup}

We conduct comprehensive experiments on three reasoning datasets to evaluate the effectiveness of Post-Completion Learning (PCL). Our evaluation covers GSM8K (8k training, 1.3k test samples), StrategyQA (2.3k samples with 90\%-10\% train-dev split), and MathQA (34k training, 3k test samples), providing diverse reasoning challenges across different scales and complexities~\citep{cobbe2021training,geva2021did,amini2019mathqa}. We will further conduct experiments on other fields like coding and open-ended generation, which is discussed in our future work.

For teacher model distillation, we use GPT-4.1 to generate post-completion training data following the PCL format. All student models are trained using the Hugging Face OpenR1 framework~\citep{openr1} with our extended \texttt{GRPOTrainer} that supports dual-track SFT training. We maintain a 1:1 ratio between the two SFT tracks (reasoning and evaluation) and a 1:1 ratio between RL and SFT training components for simplicity, training for 2 epochs with other hyperparameters following the original repository defaults.

During inference, we use the \texttt{<post-completion>} token as a stopping criterion, ensuring theoretical zero computational overhead compared to standard generation. Our evaluation focuses on accuracy as the primary metric, measuring the model's ability to produce correct final answers.

We set up a detailed experimental comparison to verify the effect of each setting, as shown in Table~\ref{tab:gsm8k_config}.

\subsection{Main Results}

\subsubsection{Performance Across Datasets}

Tables~\ref{tab:gsm8k_results}, \ref{tab:strategyqa_results}, and \ref{tab:mathqa_results} present our main experimental results across three reasoning datasets and multiple model scales. PCL demonstrates consistent and improvements over all baseline methods. 

Traditional methods typically employ \texttt{math\_verify} to extract answers through regular expression matching on the model's complete response. In contrast, our approach requires the model to strictly adhere to a predefined format and performs \textbf{exact} matching on the \texttt{<answer>} part. Consequently, our evaluation criteria are relatively more stringent, which may result in lower baseline performance metrics compared to other studies.

\textbf{GSM8K Results}: PCL achieves remarkable performance gains across all model sizes, with an average improvement of +7.81\% over SFT baselines and +3.45\% over the SFT+RL baseline. Notably, smaller models benefit more significantly, with LLaMA3.2-3B showing the largest improvement (+10.47\%), while larger models like Qwen2.5-7B still achieve substantial gains (+5.76\%). This suggests that PCL's post-completion learning paradigm is particularly effective for enhancing the reasoning capabilities of resource-constrained models.

\textbf{StrategyQA Results}: The method shows excellent consistency on this strategic reasoning dataset, achieving +6.11\% average improvement over SFT and +4.56\% over SFT+RL baselines. The uniform improvements across all model scales (Qwen2.5-7B: +5.24\%, Qwen2.5-1.5B: +6.55\%, LLaMA3.1-8B: +6.55\%) demonstrate PCL's robustness across different reasoning task types. Our method stimulates the deep reasoning ability of the model and achieves significant performance on multi-hop logical reasoning datasets.

\begin{table}[t]
\centering
\caption{Results on StrategyQA Dataset (\% Accuracy)}
\label{tab:strategyqa_results}
\footnotesize
\resizebox{0.47\textwidth}{!}{
\begin{tabular}{l@{\hspace{1em}}c@{\hspace{1em}}c@{\hspace{1em}}c@{\hspace{1em}}c}
\toprule
& \textbf{Qwen2.5} & \textbf{Qwen2.5} & \textbf{LlaMA3.1} & \\
\textbf{Method} & \textbf{(7b)} & \textbf{(1.5b)} & \textbf{(8b)} & \textbf{Average} \\
\midrule
SFT & 69.00 & 64.63 & 73.36 & 69.00 \\
SFT + RL & 69.87 & 66.38 & 75.41 & 70.55 \\
\textbf{PCL} & \textbf{74.24} & \textbf{71.18} & \textbf{79.91} & \textbf{75.11} \\
\midrule
\textbf{Improvement vs. SFT} & \textbf{+5.24} & \textbf{+6.55} & \textbf{+6.55} & \textbf{+6.11} \\
\quad \textbf{vs. SFT + RL} & \textbf{+4.37} & \textbf{+4.80} & \textbf{+4.50} & \textbf{+4.56} \\
\bottomrule
\end{tabular}
}
\vspace{0.3em}

% \footnotesize
% Best results in each column are highlighted in \textbf{bold}. PCL demonstrates consistent and substantial improvements across all model scales on the StrategyQA reasoning dataset.
\end{table}

\begin{table}[t]
\centering
\caption{Results on MathQA Dataset (\% Accuracy)}
\label{tab:mathqa_results}
\footnotesize
\resizebox{0.47\textwidth}{!}{
\begin{tabular}{l@{\hspace{1em}}c@{\hspace{1em}}c@{\hspace{1em}}c@{\hspace{1em}}c}
\toprule
& \textbf{Qwen2.5} & \textbf{Qwen2.5} & \textbf{LlaMA3.1} & \\
\textbf{Method} & \textbf{(7b)} & \textbf{(1.5b)} & \textbf{(8b)} & \textbf{Average} \\
\midrule
SFT & 72.73 & 54.71 & 54.64 & 60.69 \\
SFT + RL & 74.81 & 60.44 & 62.21 & 65.82 \\
\textbf{PCL} & \textbf{78.02} & \textbf{60.40} & \textbf{62.45} & \textbf{66.96} \\
\midrule
\textbf{Improvement vs. SFT} & \textbf{+5.29} & \textbf{+5.69} & \textbf{+7.81} & \textbf{+6.27} \\
\quad \textbf{vs. SFT + RL} & \textbf{+3.21} & \textbf{-0.04} & \textbf{+0.24} & \textbf{+1.14} \\
\bottomrule
\end{tabular}
}
\vspace{0.3em}

% \footnotesize
% Best results in each column are highlighted in \textbf{bold}. On MathQA, PCL shows strong improvements on larger models, with mixed results on smaller models, indicating dataset-specific performance characteristics.
\end{table}

\textbf{MathQA Results}: This dataset reveals more nuanced performance characteristics. While larger models show strong improvements (Qwen2.5-7B: +5.29\% vs SFT), smaller models exhibit mixed results, showing slight performance improvements or drops. We observed that the format reward curve had very obvious oscillations in the early stage of training for smaller models during training, which may be the potential reason for the final performance.

\subsubsection{Comparison with Standard Approaches}

Our results demonstrate clear advantages over traditional training paradigms. The ``RFT (Joint SFT+RL) w/ eval output'' baseline, which simply adds post-completion content without our specialized training strategy, achieves only modest gains (65.93\% vs 67.93\% for complete PCL on GSM8K). This confirms that PCL's effectiveness stems from its systematic training approach rather than merely expanding the output space.

Sequential SFT+RL approaches show improvements over SFT alone but remain consistently outperformed by PCL across all datasets. The joint training variants perform better than sequential approaches, highlighting the importance of concurrent optimization, which PCL extends through its dual-track strategy.

Besides, the ``Teacher distillation'' baseline, which uses corrected responses from teacher models for training, also achieves limited improvements (63.37\% vs 60.12\%). Compared with the performance of PCL (67.93\%), it can be demonstrated that our approach does not rely on distillation to improve performance.

\subsection{Ablation Studies}

We conduct systematic ablation studies to understand the contribution of each PCL component. Table~\ref{tab:gsm8k_results} shows results for three key ablations on GSM8K:

\textbf{Teacher Distillation}: The SFT of distilled teacher output content has a certain improvement compared to ordinary SFT (+3.25\%), which shows that directly performing ordinary post-training can also effectively improve the training effect by improving the evaluation ability of the model. However, the benefits brought by this method are lower than those of the series of methods combined with RL. These results indicate that RL can further improve the model training effect from a global perspective, and on the other hand, also show the effectiveness of our proposed method.

\textbf{Evaluation SFT Training}: Removing the evaluation-specific SFT component (SFT-E) reduces performance to 67.02\%, which confirms that explicit training on evaluation capabilities takes effect in PCL's effectiveness.

\textbf{Consistency Reward}: Eliminating the consistency reward function yields 66.08\% performance, showing a meaningful contribution. This reward helps align the model's self-assessment with ground truth, contributing to more reliable evaluation capabilities.

\subsection{Analysis and Discussion}

\subsubsection{Dataset-Specific Performance Patterns}

The varying performance across datasets provides insights into PCL's applicability. GSM8K and StrategyQA, which require multi-step logical reasoning with clear intermediate steps, show consistent improvements across all model sizes. MathQA, with its larger scale and more complex mathematical concepts, demonstrates benefits for larger models, suggesting that post-completion learning effectiveness scales with base model capabilities for highly complex reasoning tasks.

\subsubsection{Training Dynamics}

During PCL training, we observe distinct learning patterns for different reward components. Format rewards rapidly converge to high values ($\sim$1.0), indicating that models quickly learn to follow the required output structure. Accuracy rewards show more gradual improvement (from $\sim$0.6 to $\sim$0.9), reflecting the more challenging nature of learning to correctly evaluate reasoning quality. The consistency reward, which we propose to supervise the self-assessment ability, fluctuates significantly at the beginning, but maintains a score of $\sim$0.9 throughout the training process.
This suggests that PCL successfully separates the learning of structural compliance from content evaluation.

Our results establish PCL as an effective and practical approach for enhancing language model reasoning capabilities. The method's consistent improvements across diverse datasets and model scales, combined with its zero-inference overhead and systematic component contributions, demonstrate its potential as a valuable training paradigm for reasoning-intensive applications.

\section{Conclusion}

We introduce Post-Completion Learning (PCL), a novel training paradigm that systematically exploits the previously underutilized sequence space after the end-of-sequence token. Our key contributions include: (1) first systematic leveraging of post-completion space for model training, (2) a white-box reinforcement learning strategy that internalizes reward functions, and (3) a unified mixed SFT+RL framework for joint multi-objective optimization.

Comprehensive experiments across multiple datasets demonstrate that PCL consistently improves reasoning performance while maintaining zero inference overhead. The method shows broad applicability across different domains and model architectures, establishing its potential as a general training paradigm for reasoning-intensive language model applications. Our approach addresses fundamental limitations of existing methods by enabling models to develop self-evaluation capabilities during training without compromising deployment efficiency.

\section{Limitations}

Our approach has several limitations that warrant consideration. First, smaller models may have limited capacity to fully benefit from post-completion learning on highly complex datasets, as evidenced by mixed results on MathQA. This suggests that the effectiveness of PCL may be constrained by the base model's representational capacity for complex reasoning tasks.

Additionally, our current evaluation focuses primarily on mathematical and logical reasoning tasks. The generalizability of PCL to other domains such as open-ended generation, dialogue, or subjective evaluation tasks remains to be thoroughly investigated.

\section{Future Work}

Several promising directions emerge from this work that could further advance the field. We are currently extending evaluation to additional domains including code generation tasks to validate PCL's broader applicability across different types of structured reasoning problems. 

Future work could explore extending PCL to multi-turn dialogue settings where each turn incorporates post-completion evaluation, enabling more sophisticated conversational agents with built-in quality control mechanisms. Additionally, developing adaptive training mechanisms that dynamically adjust SFT and RL ratios based on training progress could improve optimization stability and efficiency, which is witnessed in our early experiments.

Another promising direction involves investigating applications to subjective tasks where self-evaluation criteria are more nuanced, such as creative writing or opinion-based reasoning. This would require developing more sophisticated reward functions and evaluation mechanisms tailored to subjective quality assessment.

Finally, exploring the integration of PCL with other recent advances in language model training, such as instruction following and constitutional AI, could lead to even more capable and reliable language models that combine multiple forms of self-improvement and quality control.

\bibliography{aaai2026}

\end{document}